\def\year{2020}\relax
\documentclass[letterpaper]{article} 
\usepackage{aaai20}  
\usepackage{times}  
\usepackage{helvet} 
\usepackage{courier}  
\usepackage[hyphens]{url}  
\usepackage{graphicx} 
\usepackage{graphicx}  
\usepackage{latexsym}
\usepackage{amsmath,amsfonts,amssymb}
\usepackage{multirow}
\usepackage{dsfont}
\usepackage[dvipsnames]{xcolor}
\newcommand{\citet}[1]{\citeauthor{#1} \shortcite{#1}}
\newcommand{\citep}{\cite}

\title{Modality-Balanced Models for Visual Dialogue}
\author{
Hyounghun Kim, 
Hao Tan, 
Mohit Bansal\\
Department of Computer Science \\
University of North Carolina at Chapel Hill\\
\{hyounghk, airsplay, mbansal\}@cs.unc.edu}
 
\begin{document}

\maketitle

\begin{abstract}
The Visual Dialog task requires a model to exploit both image and conversational context information to generate the next response to the dialogue.
However, via manual analysis, we find that a large number of conversational questions can be answered by only looking at the image without any access to the context history, while others still need the conversation context to predict the correct answers.
We demonstrate that due to this reason, previous joint-modality (history and image) models over-rely on and are more prone to memorizing the dialogue history (e.g., by extracting certain keywords or patterns in the context information), whereas image-only models are more generalizable (because they cannot memorize or extract keywords from history) and perform substantially better at the primary normalized discounted cumulative gain (NDCG) task metric which allows multiple correct answers.
Hence, this observation encourages us to explicitly maintain two models, i.e., an image-only model and an image-history joint model, and combine their complementary abilities for a more balanced multimodal model. 
We present multiple methods for this integration of the two models, via ensemble and consensus dropout fusion with shared parameters.
Empirically, our models achieve strong results on the Visual Dialog challenge 2019 (rank 3 on NDCG and high balance across metrics), and substantially outperform the winner of the Visual Dialog challenge 2018 on most metrics.
\end{abstract}

\begin{figure}[t]
\centering
  \includegraphics[width=0.95\linewidth]{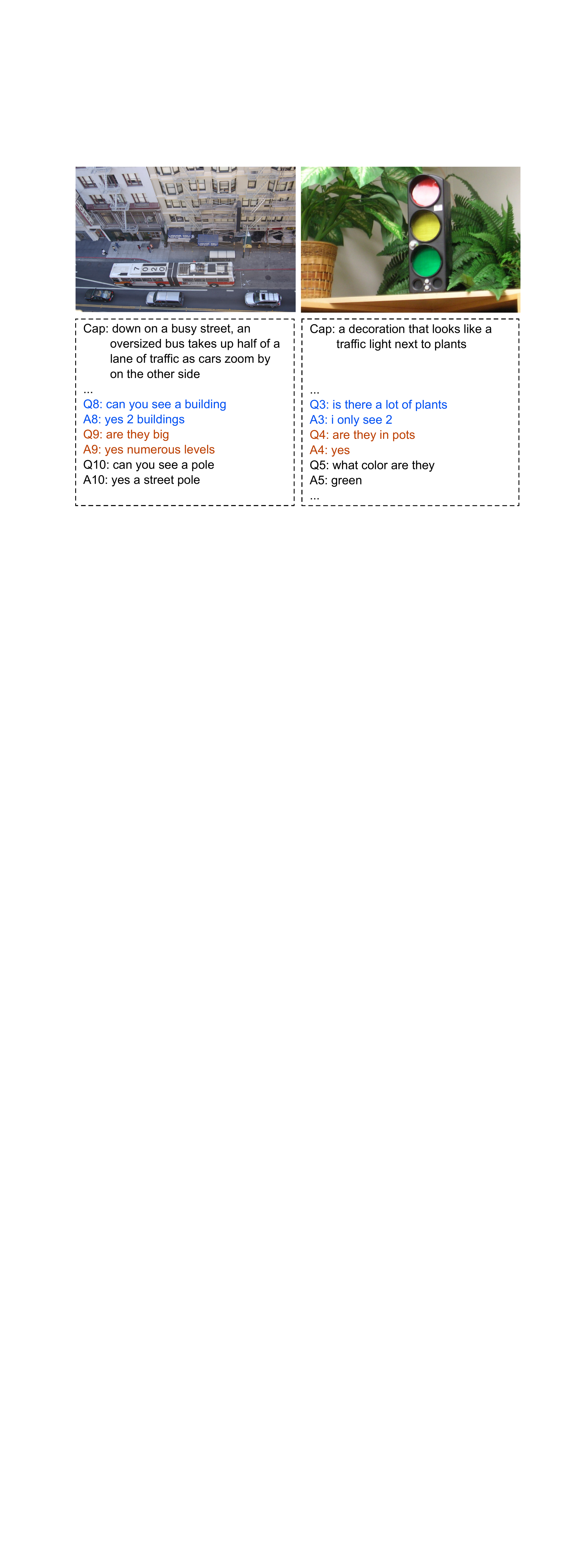}

\caption{Examples of Visual Dialog Task. Some questions only need an image to be answered (Q8-A8 and Q3-A3 pairs in blue from each example, respectively), but others need conversation history (Q9-A9 and Q4-A4 pairs in orange from each example, respectively). \label{fig:imgQA}}

\end{figure}

\section{Introduction\label{sec:intro}}

When we pursue conversations, context is important to keep the topic consistent or to answer questions which are asked by others, since most new utterances are made conditioned on related mentions or topic clues in the previous utterances in the conversation history. 
However, conversation history is not necessarily needed for all interactions, for instance, someone can change topics during a conversation and can ask a sudden new question which is not related to the context. 
This is similar to the setup in the Visual Dialog task \cite{visdial}, in which one agent (say the `asker') keeps asking questions and the other one (say the `answerer') keeps answering the questions based on an image for multiple rounds. The asker can ask a question from the conversation context. Then the answerer should answer the question by considering the conversation history as well as the image information, e.g., if the asker asks a question, ``Are they in pots?'' (Q4 in Fig.~\ref{fig:imgQA}), the answerer should find a clue in the past question-answer pairs ``Is there a lot of plants?'' - ``I only see 2.'' (Q3-A3 in Fig.~\ref{fig:imgQA}) and figure out what `they' means first to answer the question correctly. On the other hand, some questions in this task are independent of the past conversation history, e.g., ``Can you see a building?'' (Q8 in Fig.~\ref{fig:imgQA}), where the answerer does not need to look at conversation context and can answer the question only based on the image information.

We first conduct a manual investigation on the Visual Dialog dataset (VisDial) to figure out how many questions can be answered only with images and how many of them need conversation history to be answered.\footnote{We also conduct the same manual investigation to see how many questions can be answered by only looking at conversation history. It turns out that only 1\% of the questions (2 from 200 questions) can be answered. This motivates us to focus on an image-history joint model (instead of a history-only model) and merge this with an image-only model.} This investigation shows that around 80\% of the questions can be answered only with images. 
Moreover, on the model side, we verify this observation by building a model that uses only images to answer questions. As expected, this image-only model works very well on the \emph{primary} task metric of NDCG (evaluated on dense annotations which consider multiple similar answers as correct ones with similarity weights on them) without any help from the conversation history (see Table~\ref{tbl:img_ony}). However, we find that the image-only model does not get higher scores on other metrics such as mean reciprocal rank (MRR), recall@k, and mean rank (evaluated on single ground-truth answers). 
Because the image-only model does not use any conversation-history information, we hypothesize that this scoring behavior might be related to the amount of history information available, and hence we also conduct additional experiments by building an image-history joint model and train it with different lengths of history features. From these experiments, we see a tendency that a model with the less amount of history features gets a higher NDCG score (with lower values for other metrics), whereas a model with more history information has the opposite behavior.
Previously, \citet{massiceti2018visual} argued that the Visdial dataset has an answer bias such that a simple model without vision or dialogue history could achieve reasonable results. However, our motivation is different from theirs. 
The purpose of our paper is to find characteristics of existing multimodal models on the dataset (which are biased towards the language information in the dialogue history), analyze behaviors of these models on different metrics, as well as employ this analysis to build better, less biased models that achieve more balanced scores.

Since NDCG measures more of a model's generalization ability (because it allows multiple similar answers), while the other metrics measure a model's preciseness, we interpret the results of these above experiments to mean that a model with more history information tends to predict correct answers by memorizing keywords or patterns in the history while a model with less history information (i.e., the image-only model) is better at generalization by avoiding relying on such exact-match extracted information. We think that an ideal model should have more balanced behavior and scores over all the metrics rather than having higher scores only for a certain metric and such a model could be considered as the one with both preciseness and generalization.
To this end, we propose two models, an image-only and an image-history-joint model. We analyze that the answers these two models produce are complementarily good, and better at different metrics. 
Hence, we integrate these two models (image-only and image-history-joint) in two ways: consensus-dropout-fusion and ensemble. 
Our final consensus-dropout-fusion ensemble model scores strongly on both NDCG and recall metrics for the VisDial v1.0 test dataset, and these scores outperform the state-of-the-art of the Visual Dialog challenge 2018 on most metrics. Also, our model shows competitive balanced results in the Visual Dialog challenge 2019 (test-std leaderboard rank 3 based on NDCG metric and high balance across metrics).

\section{Related Work}
\paragraph{Visual Question Answering (VQA)}
Visual question answering is a task in which a machine is asked to answer a question about an image. The recent success of deep neural networks and massive data collection \cite{VQA} has made the field more active. One of the most challenging parts of the task is to ground the meaning of text on visual evidence. Co-attention \cite{lu2016hierarchical} is proposed to integrate information from different modalities (i.e., image and language) and more advanced approaches have shown good performance \cite{yu2017multi,nam2017dual,nguyen2018improved}. A bilinear approach has also been proposed to replace simple addition or concatenation approaches for fusing the two modalities \cite{gao2016compact,fukui2016multimodal,DBLP:conf/iclr/KimOLKHZ17,ben2017mutan}. In our work, we employ multi-modal factorized bilinear pooling (MFB) \cite{yu2017multiMFB} to fuse a question and image-history features.

\paragraph{Visual Dialog}
The Visual Dialog task \cite{visdial} can be seen as an extended version of the VQA task, with multiple rounds of sequential question-answer pairs as dialog history, including an image caption, which should be referred to before answering a given question. This conversation history can help a model better predict correct answers by giving direct or indirect clues for the answers, or proper context for co-reference resolution. However, having conversation history also means that a model should extract relevant information from the history and introduces another challenge to the task. Many approaches have been proposed to handle this challenge. \citet{niu2018recursive} tries to extract the clues from history recursively while \citet{wu2018you} and \citet{guo2019image} employ co-attention to fuse visual, history, and question features. In our work, we employ \citet{DBLP:conf/iclr/SeoKFH17}'s approach to fuse visual and history features before they are attended by a question. Our joint model with fused features has much information from history and we find that it is in complementary relation with our image-only model. Thus, we combine the two models to take the most appropriate information from each model to answer questions.

\section{Models}

\begin{figure*}[t]
\centering
  \includegraphics[width=2.00\columnwidth]{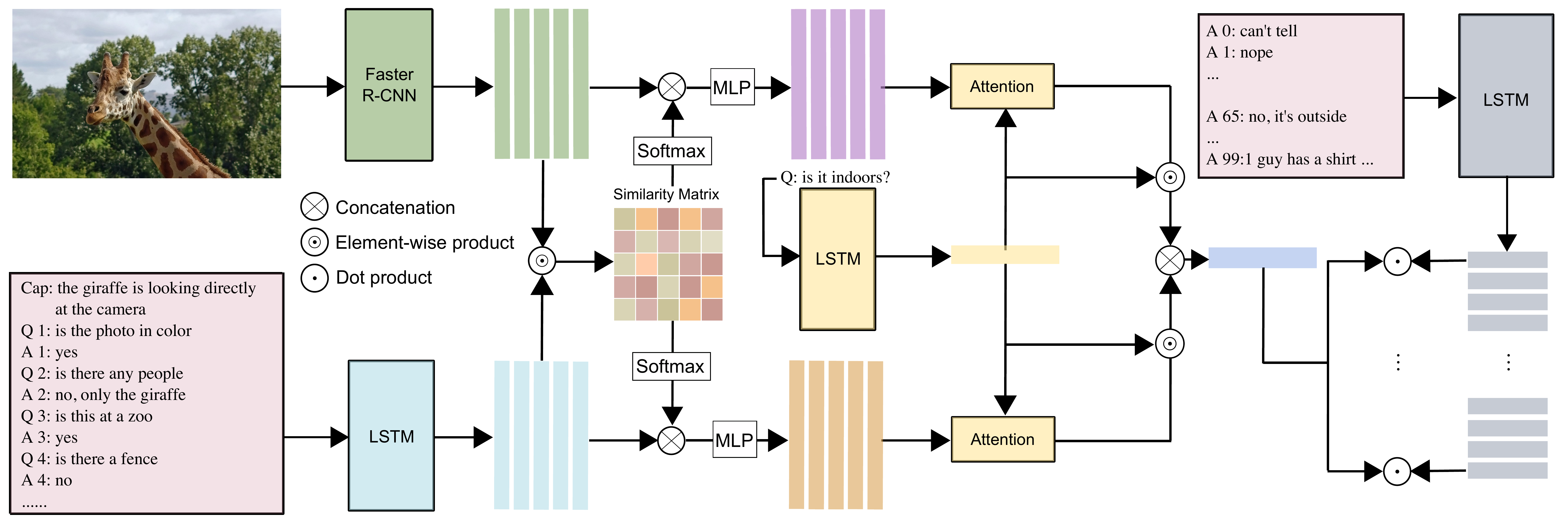}

\caption{The architecture of the image-history joint model. The visual features are obtained from Faster R-CNN and the history features are encoded via LSTM. They are fused together via the similarity matrix calculated using cross-attention. The fused features are combined with a question feature and dot products are calculated between the combined feature and candidate answers to rank the answers. \label{fig:model}}

\end{figure*}

In the Visual Dialog task \cite{visdial}, two agents interact via natural language with respect to an image. The asker keeps asking about the image given an image caption without seeing the image. 
The other agent (i.e., answerer) keeps answering the questions by viewing the image. They conduct multiple rounds of conversation accumulating question-answer pairs which are called `history' (Figure \ref{fig:imgQA}). The full history \(\textrm{HISTORY}\) consists of question-answer pairs as well as an image caption which describes the given image, such that at a current time point $t$, the previous history is \(\textrm{HISTORY}_t = \{C, (Q_{1},A_{1}), (Q_{2},A_{2}), ..., (Q_{t-1},A_{t-1}) \}\), where \(C\) is the image caption and \(Q_{t-1}\) and \(A_{t-1}\) are the question and answer at round \(t-1\), respectively. Then, given a new current time-stamp question \(Q_t\), the history \(\textrm{HISTORY}_t\), and the image, the model has to rank 100 candidate answers from the answerer's perspective.

\subsection{Features}

\noindent\textbf{Visual Features}: For visual features, we use object features which are extracted from an image by using Faster R-CNN \cite{ren2015faster}. The visual feature, \(V_{rcnn} \in \mathbb{R}^{k \times d_{v}}\), is a matrix whose rows correspond to objects, where \(k\) is the number of objects (k=36 in our experiment), \(d_{v}\) is dimension size of visual feature ($d_{v}$ = 2048 for ResNet backbone).

\noindent\textbf{Question Features}: The word sequence of a question at round \(r\), \(W_{q_{r}} = \{w_{q_{r}1}, w_{q_{r}2},..., w_{q_{r}T_{q_r}}\}\) is encoded via an LSTM-RNN \cite{hochreiter1997long},
 \begin{equation}
    h_t^{q_{r}} = \textrm{LSTM}_{q}(w_{q_{r}t}, h_{t-1}^{q_{r}})
\end{equation}
and, we take the last hidden state as a question representation: $q_{r} = h_{T_{q_{r}}}^{q_{r}}$, where \(T_{q_{r}}\) is the length of the question at round \(r\).

\noindent\textbf{History Features}: History \(H_r\) is a history feature at round \(r\) encoded from concatenation of a question and a ground truth answer, such that
\begin{equation}
\begin{split}
    W_{h_r} &= \{w_{q_{r-1}1},.., w_{q_{r-1}T_{q_{r-1}}}, w_{a_{r-1}1},.., w_{a_{r-1}T_{a_{r-1}}}\} \\
    &= \{w_{h_{r}1}, w_{h_{r}2}, ..., w_{h_{r}T_{h_{r}}}\}
\end{split}
\end{equation}
 where \(T_{a_{r-1}}\) is the length of the answer of round \(r-1\), and the length of history at round \(r\) is \(T_{h_{r}}=T_{q_{r-1}}+T_{a_{r-1}} \). The history \(H_r\) is also encoded with an LSTM,
 \begin{equation}
    h_t^{h_r}=\textrm{LSTM}_{h}(w_{h_{r}t}, h_{t-1}^{h_r})
\end{equation}
We also take the last hidden state as history representation at round \(r\): $H_r = h_{T_{h_r}}^{h_r}$. Note that the first history feature \(H_1\) comes from the image caption $C$.

\subsection{Image-Only Model}
We first build a model which only uses visual features to answer questions. We employ a state-of-the-art `bottom-up and top-down' approach from \citet{Anderson2017up-down}, in which we apply the attention mechanism over detected object features. 
We also adopt the multi-modal factorized bilinear pooling (MFB) method~\cite{yu2017multiMFB} to calculate attention weights over the visual features with respect to a question feature. 
From projected visual features and a question feature, we obtain \(z \in \mathbb{R}^{k \times d_{m}}\) by applying MFB: 
\begin{equation}
    V = \textrm{Linear}_{d_v\times d}(V_{rcnn})
\end{equation}
where \(\textrm{Linear}_{d_v\times d}\) is a linear projection which projects points from a \(d_v\)-dimension space to a \(d\)-dimension space. 
\begin{equation}
    z_{r} = \textrm{MFB}(V, q) = \sum_{i=1}^{m} ((M_iV^{\top})\odot(N_i q_{r} \cdot \mathds{1}_k^{\top}))^{\top} \
\end{equation}
where \(M\), \(N\) \(\in \mathbb{R}^{d_{m} \times d \times m}\) are trainable parameters, \(d\) is the dimension of projected visual features and a question feature, \(d_m\) is dimension of the fused feature, and \(m\) is the number of factors. \(\mathds{1}_k\) \(\in \mathbb{R}^k\) is a vector whose elements are all one. Following \citet{yu2017multiMFB}, we also apply the power normalization and \(\ell_2\) normalization to obtain \(\hat{z}_{r}\). After applying linear projection, the softmax operation is applied to get a weight vector \(\alpha\): $\alpha_{r} = \textrm{softmax}(L\hat{z}_{r}^{\top})$. We then get a visual representation vector, \(v_{r}\) by weighted summing the projected visual features: $v_{r} = \sum_{i=1}^k \alpha_{ri}V_i$, where \(L \in \mathbb{R}^{1 \times d_m }\) is trainable parameter, and \(V_i\) is the \(i\)-th row vector of visual feature matrix \(V\). The visual representation vector and a question feature vector are combined with element-wise product after linear projection. After one more linear projection, we get the final feature, \(f_{v_{r}}^{q_{r}}\) which is further used to rank answers.
\begin{equation}
    f_{v_{r}}^{q_{r}} = \textrm{fc}_f(\textrm{fc}_v(v_{r}) \odot \textrm{fc}_q(q_{r}))
\end{equation}
where \(\textrm{fc}_*\) is an fully-connected layer.

\paragraph{Answer Selection}
For each round, there are 100 candidate answers. The \(l\)-th answer at round \(r\), 
\begin{equation}
   A_{rl} = \{w_{rl1}, w_{rl2}, ... w_{rlT_{a_{rl}}}\} 
\end{equation}
 is encoded in the same way as question and history. 
 \begin{align}
    h_t^{a_{rl}} &= \textrm{LSTM}_{a}(w_{rlt}, h_{t-1}^{a_{rl}})
    \\
    a_{rl} &= h_{T_{a_{rl}}}^{a_{rl}}
\end{align}
where \(T_{a_{rl}}\) is the length of the \(l\)-th candidate answer. 
Scores for each candidate answer are calculated by dot product between fused feature \(f_{v_r}^{q_r}\) and each candidate answer representation, \(a_{rl}\): $s_{rl} = f_{v_r}^{q_r}\cdot a_{rl}$.

\begin{figure}[t]
\centering
  \includegraphics[width=.55\columnwidth]{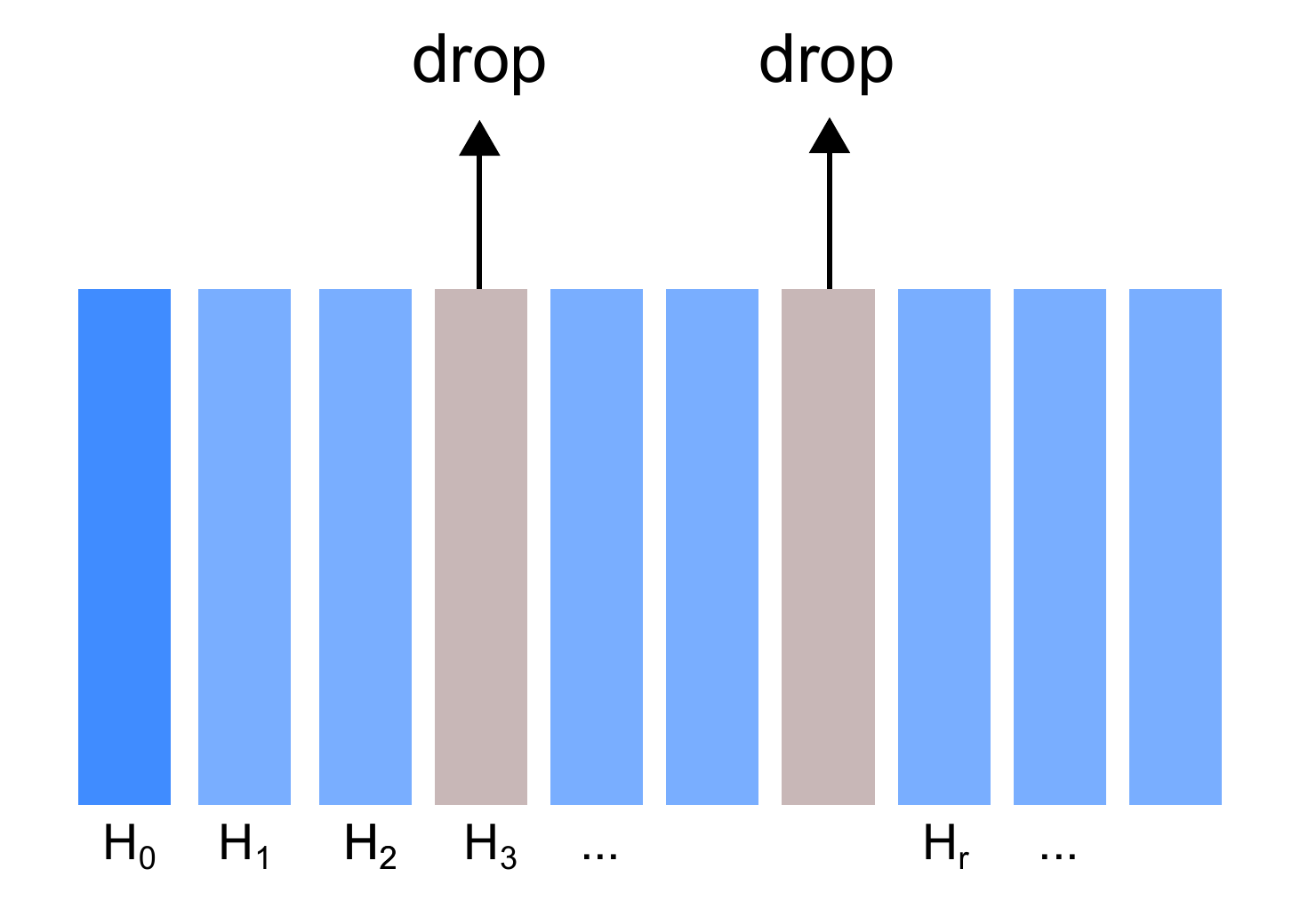}

\caption{Round Dropout: history features are dropped randomly. \(\textrm{H}_\textrm{0}\) is the image caption, \(\textrm{H}_\textrm{r}\) is the history feature at round \(r\). Dropout is not applied to the image caption feature. \label{fig:dropout}}

\end{figure}

\subsection{Image-History Joint Model}
We calculate the similarity matrix, \(S_r \in \mathbb{R}^{k \times r} \) between visual and history features following \citet{DBLP:conf/iclr/SeoKFH17}. 
\begin{equation}
    (S_{r})_{ij} = w_s^{\top}[V_i;H_j;V_i \odot H_j]
\end{equation}
where \(w_s \in \mathbb{R}^{3d}\) is trainable parameter and \(H_j\) is the \(j\)-th row vector of the history feature \(H_{1:r}\). From the similarity matrix, the new fused history representation is:
\begin{align}
    V_r^h &= \textrm{softmax}(S_r^{\top})V \\
    H_{1:r}^f &= [H_{1:r};V_r^h;H_{1:r} \odot V_r^h]
\end{align}
Similarly, the new fused visual representation is:
\begin{align}
    H_r^v &= \textrm{softmax}(S_r)H_{1:r} \\
    V_r^f &= [V;H_r^v;V \odot H_r^v]
\end{align}
These fused features are then fed to the MFB module and attended over w.r.t. a question feature, respectively, following the same process as a visual feature in the image-only model. The weighted-summed features are combined with a question feature through element-wise product and concatenated together to produce the integrated representation:
\begin{align}
    f_{v_{r}}^{q_{r}} &= \textrm{fc}_v(v_{r}^f) \odot \textrm{fc}_q(q_{r}) \\
    f_{h_{r}}^{q_{r}} &= \textrm{fc}_h(h_{r}^f) \odot \textrm{fc}_q(q_{r}) \\
    f_{v{_{r}}h{_{r}}}^{q_{r}} &= \textrm{fc}_f([f_{v_{r}}^{q_{r}};f_{h_{r}}^{q_{r}}])
\end{align}
where \(v_{r}^f\) and \(h_{r}^f\) are weighted-sum of fused features with respect to a question feature. Figure \ref{fig:model} depicts the whole process of the joint model in this section.

\begin{figure}[t]
\centering
  \includegraphics[width=.65\columnwidth]{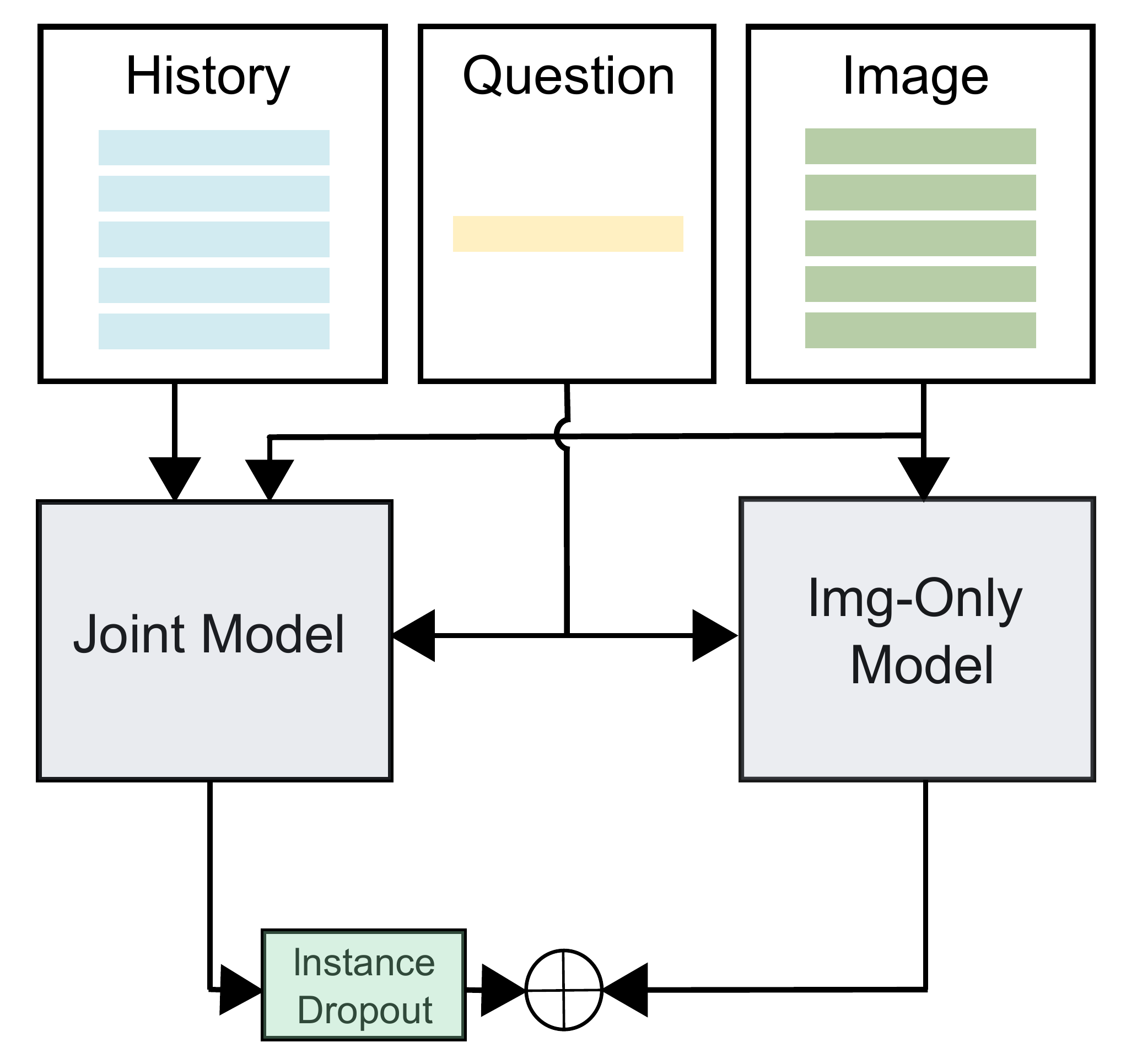}

\caption{Consensus Dropout Fusion. Logits from both image-only model and joint model are added to produce combined one. Instance dropout is applied to the logit from joint model to prevent strong coupling. The two models share many portions of parameters and are trained together.  \label{fig:latefusion}}

\end{figure}

\paragraph{Round Dropout \label{model:rd}}
To prevent the model from over-relying on history information, we propose a novel dropout approach in which some rounds of history features are dropped out (Figure \ref{fig:dropout}). To be specific, we randomly pick up to 3 rounds of history from entire history except image caption feature and throw them away.
\begin{equation}
    N_D^r = \left\{ \begin{array}{cl}
  \textrm{max}(0, N_h^r - 2) &\mbox{ if } N_h^r \leq 5 \\
    3 &\mbox{ otherwise}
    \end{array} \right.
\end{equation}
where \(N_h^r\) is number of history features at round \(r\) and \(N_D^r\) is the number of history features to drop at round \(r\).

\subsection{Combining Image-Only \& Image-History Joint Models \label{model:combine}} 
Since each of our models has different abilities, we exploit their complementary abilities together by combining them in two ways. The first is our novel consensus dropout fusion which integrates the two models in training time. The other way is to build an ensemble model from the two models at test time.

\subsubsection{Consensus Dropout Fusion \label{model:cdf}}
In order to integrate the image-only model and the image-history joint model into one model, we propose a novel integration method called consensus dropout fusion. Our consensus dropout fusion is the combination of a consensus method and an instance dropout method (Figure \ref{fig:latefusion}).

\paragraph{Consensus}
We employ a consensus method in which logits from each model are added to produce the final logit following \citet{wang2016temporal}'s approach. 
\begin{equation}
    L_{IJ} = L_{I} + L_{J}
\end{equation}
where \(L_{I}\) and \(L_{J}\) are the logit from image-only model and image-hitory joint model, respectively, and \(L_{IJ}\) is the new logit obtained by adding the two logits.

\paragraph{Instance Dropout}
To allow the image-only model to have a stronger effect producing more balanced results over all metrics, we apply dropout to instances of the logit of the joint model. To be specific, when we add two logits, we multiply \(L_{J}\) by \(I_{drop}\),

\begin{align}
    L_{J}^{drop} &= I_{drop} \odot L_{J} \\
    I_{drop} &= (\mathds{1}_{(N\times R)} \odot \xi) \cdot \mathds{1}_{d}^{\top} \\
    \xi_{i} &\sim \frac{1}{1-p}\textrm{Bernoulli}(1-p)
\end{align}
where \(\mathds{1}_{(N\times R)} \in \mathbb{R}^{(N\times R)}\) and \(\mathds{1}_{d} \in \mathbb{R}^{d}\) are all-ones vectors of \((N\times R)\) and \(d\) dimension, respectively. \(N\) is the training batch size and \(R\) is the length of rounds of the conversation history. The dropout mask, \(\xi\), is calculated following \citet{srivastava2014dropout}'s work.

\subsubsection{Ensemble}
We also integrate our 2 models via an ensemble. We train each model separately and combine them at test time. To be specific, we take logits from the pre-trained models and select the answer with the highest sum of logits.

\section{Experimental Setup}
\subsection{Dataset}
We use the VisDial v1.0 \cite{visdial} dataset to train our models, where one example has an image with its caption, 9 question-answer pairs, and follow-up questions and candidate answers for each round. At round \(r\), the caption and the previous question-answer pairs become conversational context. The whole dataset is split into 123,287\slash 2,000\slash 8,000 images for train\slash validation\slash test, respectively. Unlike the images in the train and validation sets, the images in the test set have only one follow-up question and candidate answers and their corresponding conversational context. 

\subsection{Metrics}
For evaluation, the Visual Dialog task employs four metrics. NDCG is the primary metric of the Visual Dialog Challenge which considers multiple similar answers as correct ones. The other three are MRR, recall@k, and mean rank where they only consider the rank of a single answer. Our experiments show the scores of NDCG and non-NDCG metrics from our image-only and joint models have a trade-off relationship due to their different ability (as shown in Sec.\ref{sec:reduced}) in completing Visual Dialog tasks: the image-only model has a high NDCG and low non-NDCG values while the joint model has a low NDCG and high non-NDCG values.  

\subsection{Training Details}
In our models, the size of word vectors is 300, the dimension of visual feature is 2048, and hidden size of LSTM units which are used for encoders of questions, context history, and candidate answers is 512. We employ Adam \cite{DBLP:journals/corr/KingmaB14} as the optimizer. We set the initial learning rate to 0.001 and decrease it by 0.0001 per epoch until 8th epoch and decay by 0.5 from 9th epoch on.
For round dropout, we set the maximum number of history features to be dropped to 3 and we tune the p value to 0.25 for our instance dropout in the consensus dropout fusion module. Cross-entropy is used to calculate the loss.

\section{Analysis and Results \label{sec:analy}}

\begin{table}[t]

\begin{center}
\begin{tabular}{c c c }
  \hline
   & Only Img. &  Need Hist.\\
  \hline
  \% of Questions & 81.0 \%  & 19.0 \% \\
  \hline
\end{tabular}
\end{center}

\caption{Human evaluation on questions of VisDial v1.0 val set. Percentage of questions which can be answered only from image or need help from conversation history is calculated by the manual investigation. \label{tbl:human}}

\end{table}

In this section, we first discuss how many questions are answered only from image and how many of them need image and history jointly to be answered by conducting a manual investigation. We find that a large portion of questions in the VisDial dataset can be answered by only using images. Next, to verify the observation from the manual investigation, we perform a follow-up experiment and find a trade-off relation between the amount of history features and the metric scoring trend of models. We then analyze the answers from two models (image-only and image-history joint model) and show they are in complementary relation. Lastly, we show each model can make up for the other by being combined in consensus dropout fusion or in an ensemble model.

\subsection{Human Evaluation: Is Image Alone Enough?}
We conduct a human evaluation on image, history, and question. To be specific, we randomly select 100 images (which leads to 1000 questions) from the validation set for the evaluation and count the number of questions which can be answered only with images and the number of questions which need conversation context to be answered (ground-truth answers are provided to check if the answers can be inferred given corresponding questions and images instead of providing all the 100 candidate answers). Two annotators conduct the experiment independently and questions on which both annotators mark as being able to be answered only with images are classified as only-image questions otherwise as need-history questions. The inter-annotation agreement (kappa) is 0.74.\footnote{Kappa of 0.74 is considered `substantial' agreement: https://en.wikipedia.org/wiki/Cohens\_kappa} As shown in Table \ref{tbl:human}, around 80\%\footnote{We compute statistical significance via bootstrap test \cite{efron1994introduction} and find that in 99,975 of 100K trials (i.e., p $<$ 0.0005), the percentage of only-image questions is over 75\%.} of the questions can be answered only from images. Conversely, this also implies that a model needs conversation context to better perform the task. However, as discussed in Sec.\ref{sec:intro}, using only history is not enough either (only 1\% of the questions can be answered) and thus history should be used jointly with images. Note that we consider a question with a pronoun as answerable only with an image if the pronoun can be inferred (co-reference) from the corresponding image (e.g., a question mentions `he' and the image has only one person who is a boy).

\begin{table}[t]
\begin{center}
\resizebox{0.95\columnwidth}{!}{
\begin{tabular}{c c c c c c c}
  \hline
  Models & NDCG  &  MRR & R@1  & R@5 & R@10 &  Mean \\
  \hline
  FULL & 57.81  & \textbf{64.47} & \textbf{50.87} & \textbf{81.38} & 90.03 & \textbf{4.10} \\
H-5  &  58.24	 & 64.29 & 50.61 & 81.35 & \textbf{90.22} & 4.10	\\
H-1  &  59.29 & 62.86 & 49.07 & 79.76 & 89.08 & 4.35 \\
Img-only & \textbf{61.04} & 61.25 & 47.18 & 78.43 & 88.17 & 4.61 \\
  \hline
\end{tabular}
}
\end{center}

\caption{Performance of models with the different amount of history on validation dataset of VisDial v1.0 (Round dropout is not applied to the joint model in these experiments. FULL: full image-history joint model, H-k: image-history joint model with k history, Img-only: image-only model. For H-k models we include image caption feature for a fair comparison with the full joint model). \label{tbl:img_ony}}

\end{table}

\subsection{Reduced Question-Answer Rounds \label{sec:reduced}}
We next run our joint model with various lengths of history. To be specific,  we make our joint model use only k previous history features to answer a question. As shown in Table \ref{tbl:img_ony}, there is a trade-off between the values of metrics and the number of history features. As the number of history features the joint model uses is increased, the score of NDCG is decreased while other metrics are increased. On the other hand, as the number of history features the joint model uses is decreased the score of NDCG is increased while other metrics are decreased. If we see the Visual Dialog primary task metric of NDCG as a barometer of the model's ability to generalize and the other metrics can be seen as an indicator of preciseness, this means that decreased size of history gives a model the ability of generalization at the cost of preciseness. From this tendency, the image-only model has the highest NDCG score.

\begin{table}[t]

\begin{center}
\resizebox{0.95\columnwidth}{!}{
\begin{tabular}{c c c c c}
  \hline
   & Img-Only Model & Joint Model  & Intersection & Union\\
  \hline
R@1 & 47.18   & 50.87 & 41.57 & 56.48\\
NDCG & 61.04  & 58.97 & 55.65 & 64.36 \\
  \hline
\end{tabular}
}
\end{center}

\caption{Intersection and Union of the answers from image-only model and joint model which contribute to scoring for R@1 and NDCG metrics.\label{tbl:compl}}

\end{table}

\subsection{Complementary Relation}
If the image-only model is good at NDCG, can we exploit its ability by combining it with the joint model? To figure out this possibility, we compare each answer from the image-only model and the joint model. To be specific, for R@1, we list up the correct answers from each model and count answers which are in both sets, i.e., the intersection. From the intersection, we obtain the union of the two sets. For NDCG, there is not one single correct answer. So we roughly calculate the intersection by taking minimum values between the two models' scores and averaging them. As we can see in Table \ref{tbl:compl}, the intersections do not take the entire score of either model for both metrics. This could mean image-only and joint models have room to be improved by combining them together.

\subsection{Model Combination Results}
Considering the complementary relation between image-only model and joint model, combining the two models would be a good approach to take the best from the both. So, we integrate these two models via two methods: consensus dropout fusion and ensemble (see Sec.\ref{model:combine}).

\subsubsection{Consensus Dropout Fusion Results}
As shown in Table \ref{tbl:ens}, consensus dropout fusion improves the score of NDCG by around 1.0 from the score of the joint model while still yielding comparable scores for other metrics. Unlike ensemble way, consensus dropout fusion does not require much increase in the number of model parameters.

\begin{table}[t]

\begin{center}
\resizebox{0.95\columnwidth}{!}{
\begin{tabular}{c c c c c c c}
  \hline
  Models & NDCG  &  MRR & R@1  & R@5 & R@10 &  Mean \\
  \hline
  Img-Only & 61.04 & 61.25 & 47.18 & 78.43 & 88.17 & 4.61
	 \\
Joint  & 58.97 & 64.57 & 50.87 & 81.58 & 90.30 & 4.05
	\\
	\hline
	
CDF  & 59.93 & 64.52 & 50.92 & 81.31 & 90.00 & 4.10

\\
Ensemble & 61.20 & 64.67 & 51.00 & 81.60 & 90.37 & 4.03
	\\
  \hline
\end{tabular}
}
\end{center}

\caption{Performance of the consensus dropout fusion model and the ensemble model between our image-only model and joint model on the validation dataset of VisDial v1.0 (Img-Only: image-only model, Joint: image-history joint model, CDF: consensus dropout fusion model). \label{tbl:ens}}

\end{table}

\subsubsection{Ensemble Model Results}
As also shown in Table \ref{tbl:ens}, the ensemble model seems to take the best results from each model. Specifically, the NDCG score of the ensemble model is comparable to that of the image-only model and the scores of other metrics are comparable to those of the image-history joint model. From this experiment, we can confirm that the two models are in complementary relation.

\begin{table*}[t]
\small
\begin{center}

\begin{tabular}{c l c c c c c c}
  \hline
  & Models & NDCG  &  MRR & R@1  & R@5 & R@10 & Mean \\
  \hline
  & LF \cite{visdial} & 45.31 & 55.42&  40.95 & 72.45&  82.83&  5.95 \\
  & HRE \cite{visdial} & 45.46 & 54.16 & 39.93 & 70.45 & 81.50 & 6.41 \\
  & MN \cite{visdial} & 47.50 & 55.49 & 40.98 & 72.30 & 83.30 & 5.92 \\
  & MN-att \cite{visdial} & 49.58 & 56.90 & 42.43 & 74.00 & 84.35 & 5.59 \\
  & LF-att \cite{visdial} & 49.76 & 57.07 & 42.08 & 74.83 & 85.05 & 5.41 \\
 & CorefNMN \cite{kottur2018visual} & 54.7 & 61.5 & 47.55 & 78.10 & 88.80 & 4.40 \\
 \hline
 \multirow{ 4}{*}{Visual Dialog challenge 2018} & RvA \cite{niu2018recursive} & 55.59 & 63.03 & 49.03 & 80.40 & 89.83 & 4.18 \\
 & USTC-YTH \cite{yang2019making} & 57.17 & 64.22 & 50.88 & 80.63 & 89.45 & 4.20 \\ 
  & DL-61 (single) \cite{guo2019image}& 57.32 & 62.20 & 47.90 & 80.43 & 89.95 & 4.17 \\
  & DL-61 (ensemble) \cite{guo2019image} & 57.88 & 63.42 & 49.30 & 80.77 & 90.68 & 3.97 \\
  \hline
  \multirow{ 6}{*}{Visual Dialog challenge 2019} & DAN (single) \cite{kang2019dual} & 57.59 & 63.20 & 49.63 & 79.75 & 89.35 & 4.30\\
  & DAN (ensemble) \cite{kang2019dual} & 59.36 & 64.92 & 51.28 & 81.60 & 90.88 & 3.92 \\
  & ReDAN+ (ensemble) \cite{gan2019multi} & 64.47 & 53.73 & 42.45 & 64.68 & 75.68 & 6.63 \\ 
  & MReaL--BDAI (not published) & 74.02 & 52.62 & 40.03 & 65.85 & 79.15 & 6.76 \\

  \cline{2-8}
& Our Image-Only (ensemble) & 60.16 & 61.26 & 47.15 & 78.73 & 88.48 & 4.46 
	\\
& Our Consensus Dropout Fusion (ensemble) & 59.49 & 64.40 & 50.90 & 81.18 & 90.40 & 3.99
	\\
  \hline
\end{tabular}

\end{center}

\caption{Performance comparison between our models and other models on the test-standard dataset of VisDial v1.0. We run two ensemble models each from 6 image-only models and 6 consensus dropout fusion models.  \label{tbl:compare}}

\end{table*}

\subsection{Final Visual Dialog Test Results}

For the evaluation on the test-standard dataset of VisDial v1.0, we try 6 image-only model ensemble and 6 consensus dropout fusion model ensemble. 
As shown in Table \ref{tbl:compare}, our two models show competitive results compared to the state-of-the-art on the Visual Dialog challenge 2018 (DL-61 was the winner of the Visual Dialog challenge 2018). Specifically, our image-only model shows much higher NDCG score (60.16). On the other hand, our consensus dropout fusion model shows more balanced results over all metrics while still outperforming on most evaluation metrics (NDCG, MRR, R@1, and R@5). Compared to results of the Visual Dialog challenge 2019, our models also show strong results. Although ReDAN+ \cite{gan2019multi} and MReaL--BDAI show higher NDCG scores, our consensus dropout fusion model shows more balanced results over metrics while still having a competitive NDCG score compared to DAN \cite{kang2019dual}, with rank 3 based on NDCG metric and high balance rank based on metric average.\footnote{We are model name `square' on \url{https://evalai.cloudcv.org/web/challenges/challenge-page/161/leaderboard/483}}

\paragraph{Ensemble on More Models}
We also run an ensemble model from our image-only, joint, and consensus dropout fusion models (6 of each and total 18 models) and evaluate it on the test-standard dataset of the VisDial v1.0. This model's scores (NDCG: 59.90, MRR: 64.05, R@1: 50.28, R@5: 80.95, R@10: 90.60, Mean: 4.00) are in between our image-only ensemble model and our consensus dropout fusion ensemble model, i.e., this ensemble model has a higher NDCG than the consensus dropout fusion ensemble model and higher non-NDCG scores than the image-only ensemble model. This result shows that our image-only, joint, and consensus dropout fusion models make up for each other by being combined in an ensemble model as we expected.

\section{Ablation Study}

\begin{table}[t]

\begin{center}
\resizebox{0.95\columnwidth}{!}{
\begin{tabular}{l c c c c c c}
  \hline
  Models & NDCG  &  MRR & R@1  & R@5 & R@10 &  Mean \\
  \hline
CA & 57.81 & 64.47 & 50.87 & 81.38 & 90.03 & 4.10
	\\
CA + RD  & 58.97 & 64.57 & 50.87 & 81.58 & 90.30 & 4.05
	\\
  \hline
\end{tabular}
}
\end{center}

\caption{The effect of round dropout: applying round dropout improves model's performance on NDCG by around 1.2 while also improving other metrics. (CA: cross-attention model (base model), RD: round dropout). \label{tbl:RD}}

\end{table}

\noindent\textbf{Round Dropout}: As shown in Table \ref{tbl:RD}, our round dropout (see Sec.\ref{model:rd}) improves the NDCG score by 1.2. A possible interpretation is that round dropout could help the model avoid from over-fitting to some patterns in the history features by intentionally dropping some of the features in the training session.

\begin{table}[t]

\begin{center}
\resizebox{0.95\columnwidth}{!}{
\begin{tabular}{l c c c c c c}
  \hline
  Models & NDCG  &  MRR & R@1  & R@5 & R@10 &  Mean \\
  \hline
CDF (p=0.00) & 59.40 & 64.61 & 51.01 & 81.73 & 90.30 & 4.06 
	\\
CDF (p=0.15) &  59.49 & 64.64 & 50.94 & 81.63 & 90.07 & 4.07
	\\
CDF (p=0.25) & 59.93 & 64.52 & 50.92 & 81.31 & 90.00 & 4.10
\\
CDF (p=0.35) & 60.11 & 64.21 & 50.56 & 81.20 & 89.84 & 4.15
\\
  \hline
\end{tabular}
}
\end{center}

\caption{Consensus dropout fusion and different dropout rates. With different dropout rates, consensus dropout fusion model yields different scores of all metrics. (CDF: consensus dropout fusion model).\label{tbl:latefusion_rate}}

\end{table}

\noindent\textbf{Consensus Dropout Fusion and Dropout Rate}: We run our consensus dropout fusion model (see Sec.\ref{model:cdf}) with different  instance dropout rates to figure out how the dropout rates affect the performance of the model. As shown in Table.\ref{tbl:latefusion_rate}, as the dropout rate increases the NDCG score is also increased while scores of non-NDCG metrics are decreased. By changing the dropout rate, we can modulate the influence of each model (image-only and joint models) over the combined model. We choose a value of 0.25 for the dropout rate since it yields more balanced scores over all metrics.

\begin{table}[t]

\begin{center}
\resizebox{0.95\columnwidth}{!}{
\begin{tabular}{c c c c c c c}
  \hline
  Models & NDCG  &  MRR & R@1  & R@5 & R@10 &  Mean \\
  \hline
  Img+Img & 61.97  & 62.24 & 48.20 & 79.49 & 88.83 & 4.41	 \\
 Joint+Joint   &  59.84	 & 65.60 & 52.06 & 82.46 & 90.87 & 3.88\\
  \hline
   
Img+Joint  &  61.50 & 65.04 & 51.38 & 81.93 & 90.45 & 3.96\\
  \hline
\end{tabular}
}
\end{center}
\caption{Performance of ensemble models with different combinations. Img+Img model (3 Img models) has highest value of NDCG while Joint+Joint (3 Joint models) model highest values for other metrics. Img+Joint model (3 Img + 3 Joint models) has more balanced results (Img: image-only model, Joint: image-history joint model).  \label{tbl:ensemble}}
\end{table}

\noindent\textbf{Ensemble Combination}:
We try different combinations from image-only and 
joint models to build ensemble models. The total number of models amounts to 3, i.e., image-only + image-only (I+I), joint + joint (J+J), and image-only + joint (I+J) ensemble models. As shown in Table \ref{tbl:ensemble}, scores of the I+J ensemble model are comparable to same-kind ensemble models (I+I and J+J). To be specific, for the NDCG metric, the I+J model outperforms the J+J model, while, for other metrics (MRR, recall@k, and mean rank), the I+J model outperforms the I+I model. This might imply that the balanced scores (i.e., high scores over all metrics) of the I+J model are from the complementary relation between image-only and image-history joint model.

\noindent\textbf{Output Examples}: Due to space constraints and no supplementary allowed in AAAI rules, we provide detailed examples in this arxiv version's appendix, showing the coreference and memorization phenomena of the joint image-history model as well as the image-only model's example outputs on image-only questions. Examples of only-image questions, and the ranking lists of the image-history joint and image-only models are also provided.

\section{Conclusion}
We first showed that current multimodal models on the Visual Dialog task over-rely on the dialogue history, and relatedly, image-only and image-history joint models achieve complementary performance gains. Hence, to balance the best abilities from each model, we proposed two ways of combining them: consensus dropout fusion and ensemble. Our consensus dropout fusion and ensemble model achieve strong ranks on multiple leaderboards. Specifically, the models show higher scores than the state-of-the-art results of the Visual Dialog challenge 2018 and more balanced scores than highest ranked results of the Visual Dialog challenge 2019.
Given the characteristics of the dataset and current model behaviors, a potential future direction is to combine the power of the two models dynamically, e.g., learn to select a proper model based on the question type.

\section*{Acknowledgments}
We thank the reviewers for their helpful comments. This work was supported by NSF Award \#1840131, ARO-YIP Award \#W911NF-18-1-0336, and faculty awards from Google, Facebook, Bloomberg, and Salesforce. The views, opinions, and/or findings contained in this article are those of the authors and should not be interpreted as representing the official views or policies, either expressed or implied, of the funding agency.

\bibliography{aaai20.bib}
\bibliographystyle{aaai.bst}

\appendix
\section*{Appendices}
\begin{figure*}[t]
\centering
  \includegraphics[width=1.80\columnwidth]{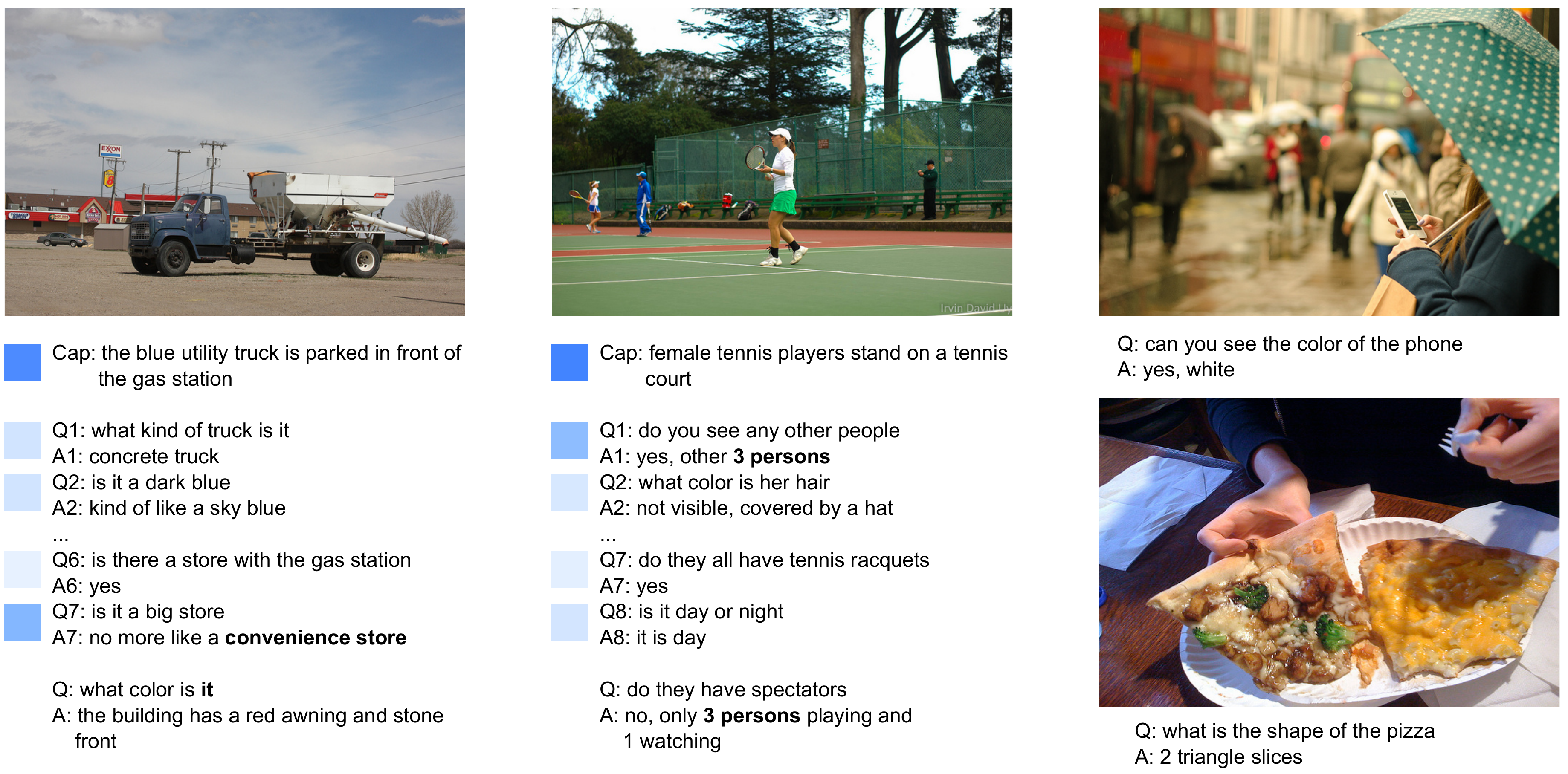}
\caption{
Coreference and memorization examples of the image-history joint model (a darker blue square indicates a higher score and a lighter blue square indicates a lower score): the left example shows that the model attends to the last QA pair to resolve the coreference (i.e., `it' to `convenience store'), and the middle example shows that the model might memorize keywords/phrases to answer questions (`3 persons'). Note that attention scores for captions are always high since they have more general information than others. On the right, we show two examples for answer prediction of the image-only model.
\label{fig:visual}}
\end{figure*}

\section{Example Analysis}
\paragraph{Coreference}
In the left example of Fig. \ref{fig:visual}, the question is ``What color is it?''. To answer this question, the model needs to know the meaning of `it'. The model attends to the last QA pair ``Is it a big store? -- No more like a convenience store.'' to figure out that `it' means 'convenience store', and gives the answer ``The building has a red awning and stone front''. Note that attention scores for captions are always high since they have more general information than others.

\paragraph{Memorization}
In the middle example of Fig. \ref{fig:visual}, the question is ``Do they have spectators?''. The model attends to the first QA pair ``Do you see any other people? -- Yes, other 3 persons.'' to give the answer ``No, only 3 persons playing and 1 watching''. However, the phrase `3 persons' from each sentence (the answer of first QA pair and the model's answer to the current question) have different meanings, and it could be seen as that the model might memorize a keyword/phrase in the history to answer questions.

\paragraph{Image-Only Model on Only-Image Question}
In the two examples in the right column of Fig. \ref{fig:visual}, the question is an only-image question and the image-only model can answer these correctly.

\begin{figure*}[t]
\centering
  \includegraphics[width=1.90\columnwidth]{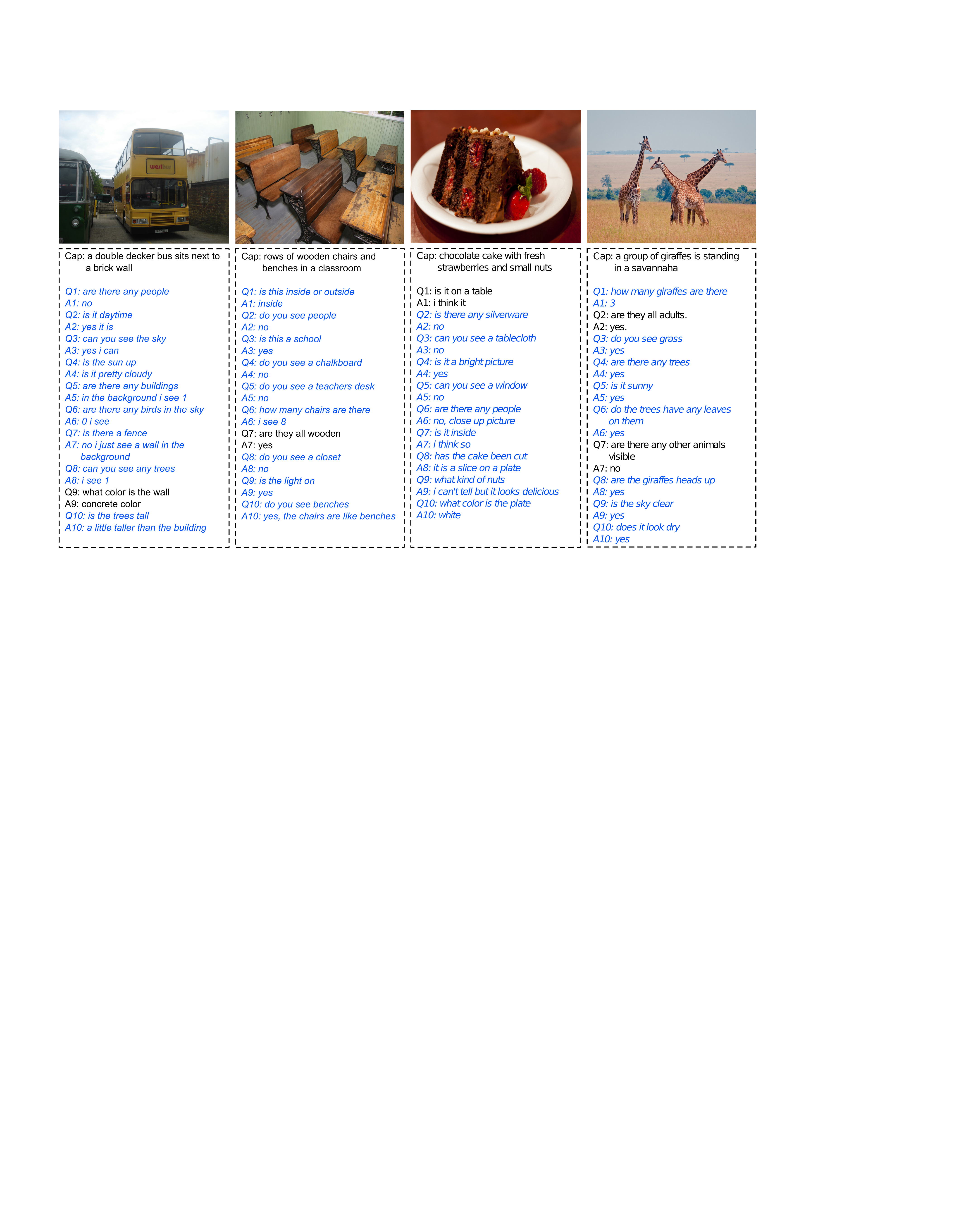}
\caption{Examples of only-image questions in blue+italics which only need an image to be answered. \label{fig:imgOnlyQ}}
\end{figure*}

\begin{figure*}[t]
\centering
  \includegraphics[width=1.50\columnwidth]{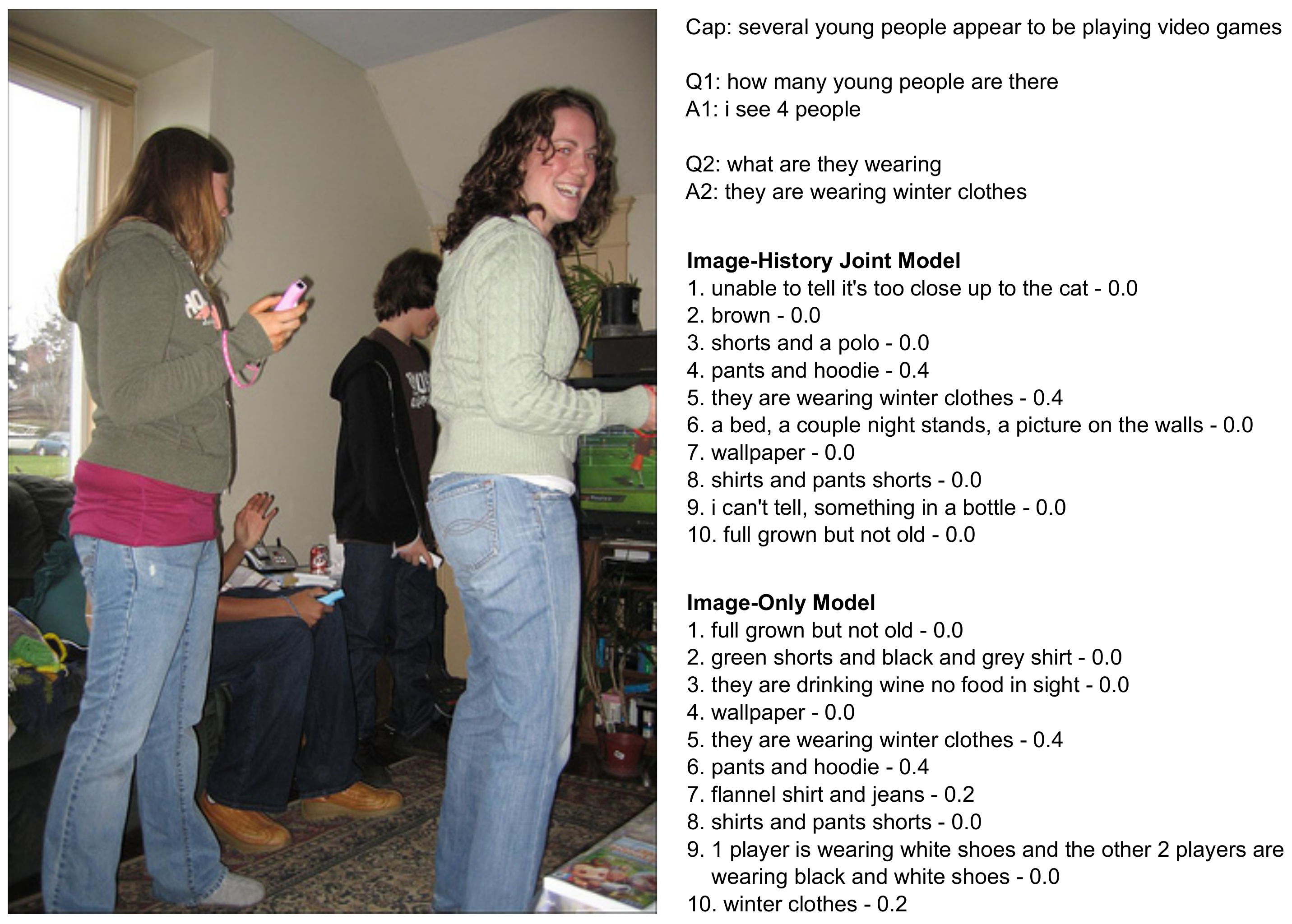}
\caption{An example of the ranking list of the image-history joint and image-only model (the numbers next to the answers are the scores which indicate the relevance of the corresponding answers to the question). \label{fig:rank}}
\end{figure*}

\section{Only-Image Question Examples}
Figure \ref{fig:imgOnlyQ} shows examples of questions that only require images to be answered. A large number of questions can be answered without looking at history information.

\section{Ranking Lists}
As shown in Fig. \ref{fig:rank}, the exact answer is the same rank in the joint and the image-only models' ranking lists, but the image-only model has a larger number of relevant answers in its list.

\end{document}